\documentclass[letterpaper,journal]{IEEEtran}
\usepackage{amsmath,amsfonts}
\usepackage{algorithmic}
\usepackage{algorithm}
\usepackage{array}
\usepackage{booktabs}
\usepackage[caption=false,font=normalsize,labelfont=sf,textfont=sf]{subfig}
\usepackage{textcomp}
\usepackage{stfloats}
\usepackage{placeins}
\usepackage{url}
\usepackage{verbatim}
\usepackage{graphicx}
\usepackage{multirow}
\usepackage{cite}
\usepackage{arydshln}
\usepackage[hidelinks]{hyperref}
\hypersetup{
  pdftitle={HiPFuse: Real-Time LiDAR Semantic Segmentation for Robotic Perception on Edge Devices},
  pdfauthor={Haiyang Wu, George Vosselman, Ville Lehtola},
  pdfsubject={Preprint},
  pdfkeywords={LiDAR semantic segmentation, point clouds, real-time perception, robotic perception}
}

\begin{document}

\title{HiPFuse: Real-Time LiDAR Semantic Segmentation for Robotic Perception on Edge Devices}

\author{Haiyang Wu$^{1}$, George Vosselman$^{1}$, and Ville Lehtola$^{1}$%
\thanks{$^{1}$Haiyang Wu, George Vosselman, and Ville Lehtola are with the Department of Geoinformatics, Faculty of Geo-Information Science and Earth Observation (ITC), University of Twente, 7522 NB Enschede, The Netherlands.\\
{\ttfamily\footnotesize haiyang.wu@utwente.nl; george.vosselman@utwente.nl; v.v.lehtola@utwente.nl}}}

\markboth{}{}

\maketitle

\begin{abstract}
Real-time robotic perception requires accurate LiDAR semantics under limited resources. Bird's-eye-view (BEV) aggregation is efficient but can mix differently labeled surfaces in columns. We present HiPFuse, a point--BEV fusion framework that exchanges point and height-partitioned BEV features across horizontal scales, combining point-wise detail with 2D spatial context from one LiDAR scan. Fixed height assignments preserve separate vertical aggregation groups, with auxiliary supervision of initial BEV features using targets from the same 3D cells. A dual-scale BEV context module adds broader context from a coarser grid. On SemanticKITTI and MCD, HiPFuse achieves 63.7\% and 62.5\% mIoU at 14.2 and 22.4~FPS on Jetson AGX Orin, respectively, exceeding the 10~Hz scan rate. Separate training and evaluation show indoor applicability. Replacing only the baseline semantic front end with HiPFuse in fixed robotic systems reduces odometry absolute trajectory error RMSE by 10.6\% on average across two outdoor sequences and increases accepted-loop recall by 8.7 percentage points on SemanticKITTI. Qualitative semantic maps show more continuous road labeling. Our code is available at \url{https://github.com/haiyang2022/HiPFuse}.
\end{abstract}

\begin{IEEEkeywords}
LiDAR semantic segmentation, point clouds, real-time perception, robotic perception.
\end{IEEEkeywords}

\section{Introduction}

Beyond scene understanding, semantic information from LiDAR scans can support state estimation tasks such as LiDAR--inertial odometry by informing how observations are selected, associated, and modeled~\cite{chen2019sumapp,wu2025gaussian}. Semantic cues also support loop retrieval~\cite{wang2024sglc} and semantic mapping~\cite{wang2025sgslam}, allowing point-wise semantic predictions to serve several components of a robotic perception pipeline. The benefit of these predictions depends on both their quality and the cost of producing them, since semantic segmentation must share computational resources with the downstream algorithms, especially on edge devices.

\suppressfloats[t]
\begin{figure}[t]
\centering
\includegraphics[width=\columnwidth]{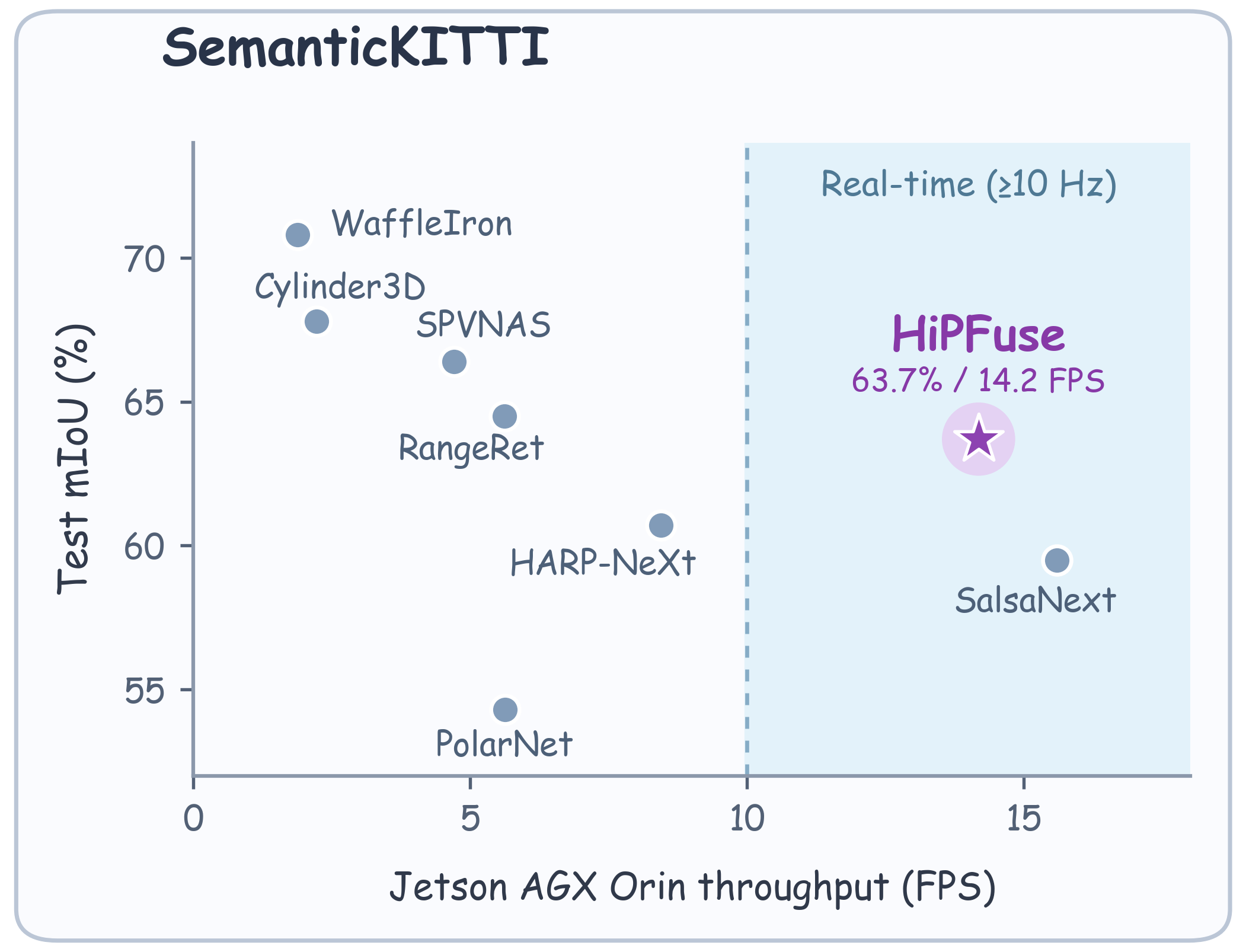}
\caption{SemanticKITTI test accuracy versus measured Jetson AGX Orin throughput (Tables~\ref{tab:semantickitti_test} and~\ref{tab:efficiency_comparison}). Shading indicates throughput meeting the 10~Hz scan rate under the measured settings.}
\label{fig:semantickitti_tradeoff}
\end{figure}

In an online perception pipeline, this calls for a semantic front end that keeps pace with incoming scans within a limited computational budget. Projecting point cloud features onto 2D grids enables efficient convolutional processing, as exemplified by the range-image networks RangeNet++~\cite{milioto2019rangenet} and SalsaNext~\cite{cortinhal2020salsanext}. However, range-image projection introduces distance-dependent variations in object scale~\cite{sirohi2022efficientlps}. A bird's-eye-view (BEV) grid instead uses fixed-size horizontal cells, providing a consistent scale for spatial processing without requiring a regular arrangement of scan lines.

This leads to a follow-up problem: surfaces at different heights within a BEV column may belong to different semantic classes, yet methods such as PolarNet~\cite{zhang2020polarnet} combine all observations in the column into one descriptor before 2D spatial processing.

The vertical axis need not be collapsed, however: height partitioning keeps separate descriptors for different height intervals within each column. Related ideas of retaining height information in BEV channels have been used in point cloud registration and LiDAR perception~\cite{li2023unified,ye2023lidarmultinet}. For segmentation, the challenge is to learn features that distinguish surfaces with different labels within the same BEV column.

A point branch offers a complementary way to preserve local detail, as demonstrated by HARP-NeXt~\cite{abou2025harp} and sparse point--voxel convolution~\cite{tang2020searching}. Keeping individual point descriptors alongside grid context does not itself separate height partitions during BEV aggregation. We bring these ideas together in HiPFuse, a point--BEV fusion framework for semantic segmentation from a single LiDAR scan. Within this framework, point features and height-partitioned BEV context are learned jointly across scales, preserving separate height groups during aggregation and individual point features during fusion.

Even with height partitioning and point-wise detail, similar local surfaces can belong to different semantic classes. Their interpretation can therefore benefit from surrounding structures and spatial relationships. To gather this context at limited additional cost, we further introduce a dual-scale BEV context (DSC) module that processes a coarser spatial grid and fuses the resulting context with features at the current resolution.

To assess HiPFuse as a semantic front end for resource-constrained robotic perception, we evaluate its accuracy and throughput on the outdoor benchmarks SemanticKITTI~\cite{behley2019semantickitti} and MCD~\cite{nguyen2024mcd}. Our method achieves a favorable accuracy--throughput trade-off relative to the evaluated baselines, with scan-rate segmentation on NVIDIA Jetson AGX Orin (Fig.~\ref{fig:semantickitti_tradeoff}). Separate training and evaluation on Semantic-ITC~\cite{wu2026semanticitc} examine indoor applicability, with segmentation throughput also exceeding the scan rate.

To determine whether this segmentation performance translates into benefits for downstream robotic perception, we replace the semantic inputs to Hybrid-VoxelMap~\cite{wu2025gaussian} odometry, SGLC~\cite{wang2024sglc} loop retrieval, and SG-SLAM~\cite{wang2025sgslam} semantic mapping while keeping the downstream algorithms and settings fixed. Compared with semantic inputs from SalsaNext~\cite{cortinhal2020salsanext} and HARP-NeXt~\cite{abou2025harp}, HiPFuse predictions reduce outdoor odometry errors and increase loop retrieval recall. Qualitative map comparisons also show more continuous road labeling.

The main contributions are as follows:
\begin{itemize}
\item We introduce HiPFuse, a point--BEV fusion framework for real-time LiDAR semantic segmentation whose integrated design combines height partitioning along the LiDAR $z$-axis, point-wise detail, and 2D spatial context learning.
\item We expand the context available for semantic interpretation with a dual-scale BEV module, limiting the added spatial computation by extracting broader context on a coarser grid.
\item We demonstrate a favorable accuracy--efficiency balance with scan-rate segmentation throughput on an edge device, and show that HiPFuse, requiring only the current scan, can be plugged in as the semantic front end of existing systems and benefits multiple robotic perception tasks while keeping their algorithms and settings fixed.
\end{itemize}

\section{Related Work}

\subsection{Efficient LiDAR Semantic Segmentation}

Efficient LiDAR segmentation organizes feature aggregation over point neighborhoods, 3D voxel cells, or projected 2D grids, with different consequences for computational cost and geometric detail. RandLA-Net~\cite{hu2020randla} operates on point neighborhoods, combining random sampling with local feature aggregation to retain information after subsampling. Cylinder3D~\cite{zhu2021cylindrical} organizes points in cylindrical cells adapted to outdoor scan density and applies asymmetrical 3D convolutions. RangeNet++~\cite{milioto2019rangenet} instead performs 2D processing on spherical range images and uses range-based postprocessing to recover point labels affected by projection and discretization.

Among projected representations, BEV grids group observations by horizontal location, making the treatment of height central to descriptor construction. PolarNet~\cite{zhang2020polarnet} pools learned point features into one descriptor per polar column for 2D processing. Its output channels encode separate class scores for each height bin and are reshaped into a voxel prediction grid. Layered alternatives retain separate height contributions within the BEV channels: VolMap~\cite{radi2019volmap} stacks BEV intensity maps from different LiDAR layers, while LidarMultiNet~\cite{ye2023lidarmultinet} concatenates height-indexed features from a downsampled voxel grid to provide BEV context within a sparse 3D encoder--decoder.

Grid organization determines which points are aggregated together, but the resulting descriptors can obscure differences needed for point-wise semantic prediction. Hybrid representations retain individual point features alongside grid context to preserve these distinctions. Sparse point--voxel convolution~\cite{tang2020searching} combines a high-resolution point stream with sparse voxel convolutions, while RPVNet~\cite{xu2021rpvnet} exchanges features among range, point, and voxel representations through gated fusion. For 2D spatial processing, WaffleIron~\cite{puy2023using} repeatedly projects point features onto Cartesian planes and lifts the processed features back to points. HARP-NeXt~\cite{abou2025harp} couples range-image and point features through hierarchical fusion, using pixel groups for local encoding and aggregation together with auxiliary range-image supervision.

Across these representation choices, context extraction must account for both spatial scale and computational cost. EfficientLPS~\cite{sirohi2022efficientlps} uses range-aware feature fusion and a distance-dependent semantic head to handle object-scale variation in LiDAR panoptic segmentation. SalsaNext~\cite{cortinhal2020salsanext} expands receptive fields using dilated convolutions without the parameter cost of corresponding dense kernels, while MINet~\cite{li2022minet} allocates more complex processing to lower-resolution paths.

These designs expose a common tension: grid aggregation supports compact spatial processing but can mix observations requiring different semantic labels. We address this tension through a point--BEV fusion framework that jointly learns height-partitioned BEV features and individual point features across scales, retaining both height and point distinctions within 2D spatial processing.

\subsection{Semantics for Robotic Perception}

For pose estimation, semantic information can guide observation selection and the construction of geometric constraints for registration. SuMa++~\cite{chen2019sumapp} uses semantic consistency to filter dynamic surfels and semantic information to weigh scan-matching residuals. Hybrid-VoxelMap~\cite{wu2025gaussian} uses semantic super-labels to select planar or Gaussian voxel models and combines their residuals for LiDAR--inertial odometry.

Beyond local scan matching, the spatial arrangement of semantic observations can help identify revisited places. Semantic Scan Context~\cite{li2021ssc} encodes categories in a polar descriptor after scan alignment. SGLC~\cite{wang2024sglc} combines foreground semantic graphs with background geometry for candidate retrieval, verification, and loop pose estimation.

Alongside their use in pose estimation and loop retrieval, semantic observations can be accumulated across scans to build persistent maps. Bayesian spatial kernel smoothing~\cite{gan2020bayesian} exploits spatial correlations to estimate semantic occupancy, while the semantic extension of PIN-SLAM~\cite{pan2024pinslam} learns a continuous semantic field from point-wise labels. SlideSLAM~\cite{liu2025slideslam} uses segmented objects as landmarks and jointly optimizes their models and robot trajectories. SG-SLAM~\cite{wang2025sgslam} maintains both semantic point cloud and graph maps, using instance tracking to associate graph nodes across scans.

We therefore evaluate semantic predictions through the odometry estimates, loop retrieval results, and semantic maps they help produce. Keeping downstream algorithms and settings fixed, we show that replacing only the semantic input benefits these perception tasks.

\section{Methodology}
\label{sec:Methodology}

\begin{figure*}[t]
\centering
\includegraphics[width=\textwidth]{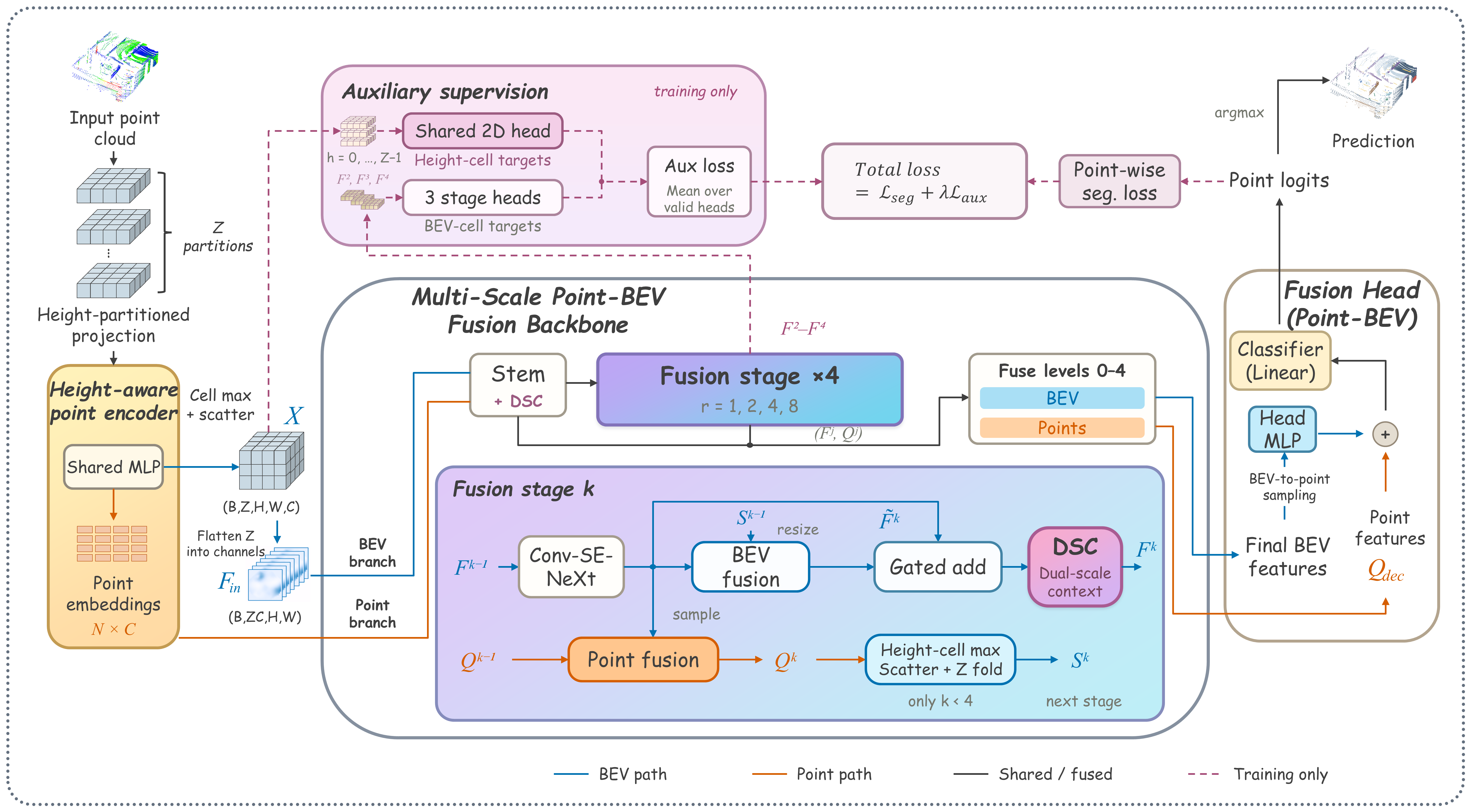}
\caption{Overview of the HiPFuse architecture.}
\label{fig:overall}
\end{figure*}

\subsection{Overview}
\label{sec:overview}

Fig.~\ref{fig:overall} illustrates the overall HiPFuse pipeline. From a single LiDAR scan, HiPFuse encodes individual points and aggregates their features in 3D cells to form a height-partitioned BEV representation. The point and BEV branches then exchange features across scales: 2D processing provides BEV context to individual points, and the updated point features are aggregated back into BEV. Throughout this process, the point branch retains individual descriptors, while BEV aggregation keeps the height assignments fixed as the horizontal grid becomes coarser. Features from multiple levels of both branches are finally combined for point-wise semantic prediction.

We adopt the Conv-SE-NeXt blocks, residual attention fusion, and fusion head from HARP-NeXt~\cite{abou2025harp}.

\subsection{Height-Partitioned Projection and Encoding}
\label{sec:projection}

\subsubsection{\textbf{BEV Grid and Height Partitions}}
Each input point has coordinates $\mathbf{p}_i=(x_i,y_i,z_i)^{\mathsf T}\in\mathbb{R}^{3}$ and intensity $\rho_i$. We divide a predefined region in the LiDAR coordinate system into a $W\times H$ BEV grid and $Z$ uniform height partitions, forming 3D cells of size $s_x\times s_y\times s_z$. Each new scan is partitioned directly along the LiDAR $z$-axis, without gravity alignment or leveling.

Let $x_{\min}$, $y_{\min}$, and $z_{\min}$ denote the region's lower coordinate bounds. Each point within this region is assigned the cell indices
\begin{equation}
(u_i,v_i,h_i)=\left(
\left\lfloor\tfrac{x_i-x_{\min}}{s_x}\right\rfloor,
\left\lfloor\tfrac{y_i-y_{\min}}{s_y}\right\rfloor,
\left\lfloor\tfrac{z_i-z_{\min}}{s_z}\right\rfloor
\right),
\label{eq:grid_index}
\end{equation}
where $u_i\in\{0,\dots,W-1\}$, $v_i\in\{0,\dots,H-1\}$, and $h_i\in\{0,\dots,Z-1\}$.

\subsubsection{\textbf{Height-Aware Point Encoder}}
\label{sec:encoding}

Using these cell assignments, we encode each point relative to its 3D cell. For a batch of $B$ scans containing $N$ retained points, let $\mathcal{V}_i=(b_i,h_i,v_i,u_i)$ identify point $i$'s cell, where $b_i$ is the scan index. The cell's mean point coordinate is $\bar{\mathbf{p}}_{\mathcal{V}_i}$, and its geometric center is $\mathbf{c}_{\mathcal{V}_i}$.
We concatenate each point's coordinates and intensity with its coordinate offsets from the cell mean and cell center, using brackets to denote concatenation. A shared point-wise multilayer perceptron (MLP) $\psi$ with Linear--BatchNorm--ReLU layers maps this input to a $C$-dimensional point descriptor $\mathbf{e}_i$:
\begin{equation}
\mathbf{e}_i =
\psi
\left(
\left[
\mathbf{p}_i,\rho_i,\,
\mathbf{p}_i-\bar{\mathbf{p}}_{\mathcal{V}_i},\,
\mathbf{p}_i-\mathbf{c}_{\mathcal{V}_i}
\right]
\right).
\label{eq:point_encoder}
\end{equation}
The mean-relative offset describes local point geometry, whereas the center-relative offset locates the point within its 3D cell.

\subsubsection{\textbf{Cell Max Pooling and Scatter}}
The encoded point features $\mathbf{e}_i$ in each 3D cell are aggregated by channel-wise max pooling. We scatter the pooled descriptors to their cell indices and set empty cells to zero, giving $\mathbf{X}\in\mathbb{R}^{B\times Z\times H\times W\times C}$.

At each horizontal cell, the $Z$ height partitions each contribute $C$ features. We arrange these $ZC$ values as channels of a 2D BEV feature map, yielding the input $\mathbf{F}_{\mathrm{in}}\in\mathbb{R}^{B\times ZC\times H\times W}$ for 2D convolution. The individual descriptors $\mathbf{e}_i$ are retained for the point branch.

\subsection{Multi-Scale Point--BEV Fusion Backbone}
\label{sec:coupling}

The backbone takes the BEV input $\mathbf{F}_{\mathrm{in}}$ and encoded point features $\{\mathbf{e}_i\}_{i=1}^{N}$ from the preceding encoder. An initial feature-processing block, referred to as the stem, initializes the point and BEV branches; four fusion stages then update them at multiple BEV resolutions. The backbone then combines these levels within each branch to provide the inputs to the fusion head. We denote the output BEV map and the set of point descriptors at level $k$ by $\mathbf{F}^{k}$ and $\mathbf{Q}^{k}=\{\mathbf{q}_i^{k}\}_{i=1}^{N}$, respectively, with $k=0$ for the stem and $k=1,\dots,4$ for the fusion stages.

\subsubsection{\textbf{Stem}}
The stem first applies three successive $3\times3$ convolutions to $\mathbf{F}_{\mathrm{in}}$. This produces the intermediate BEV map $\widetilde{\mathbf{F}}^{0}$ at resolution $H\times W$, before point--BEV fusion. For each point, we sample this map at its horizontal cell, concatenate the sampled features with $\mathbf{e}_i$, and apply a shared point-wise MLP $\phi_0$ to obtain the updated descriptor $\mathbf{q}_i^{0}$ for point $i$.

We then aggregate these descriptors, collected in $\mathbf{Q}^{0}$, into a BEV tensor $\mathbf{S}^{0}$ using the same cell max pooling, scatter, and channel arrangement as for $\mathbf{F}_{\mathrm{in}}$. A 2D convolutional fusion layer $\gamma_0$ projects the concatenation of $\mathbf{S}^{0}$ and $\widetilde{\mathbf{F}}^{0}$, and the DSC module $\mathcal{D}_0$ (Sec.~\ref{sec:dual_scale}) adds broader context to produce $\mathbf{F}^{0}$. The stem convolutions, point update MLPs $\phi_k$, and BEV fusion layers $\gamma_k$ all use batch normalization and Hardswish~\cite{howard2019mobilenetv3}.

\subsubsection{\textbf{Fusion Stages}}
Each stage $k$ receives the preceding BEV features $\mathbf{F}^{k-1}$, point descriptors $\mathbf{Q}^{k-1}$, and their BEV aggregate $\mathbf{S}^{k-1}$. Stage 1 retains the stem resolution, while stages 2--4 each halve both BEV dimensions, giving cumulative horizontal strides $(r_1,r_2,r_3,r_4)=(1,2,4,8)$. We choose $H$ and $W$ to be divisible by 8. The point branch retains a descriptor for each of the $N$ points at every stage.

A stack of Conv-SE-NeXt blocks $\mathcal{B}_k$ processes $\mathbf{F}^{k-1}$ through spatial convolution, channel mixing, and channel attention to produce the intermediate BEV features $\widetilde{\mathbf{F}}^{k}$ at resolution $(H/r_k)\times(W/r_k)$. These features are shared by the point and BEV updates.

To update the point descriptors, we sample $\widetilde{\mathbf{F}}^{k}$ at each point's horizontal cell. For a BEV feature map $\mathbf{F}$ at cumulative stride $r$, the sampled vector for point $i$ is denoted by $\mathcal{G}_r(\mathbf{F})_i$:
\begin{equation}
\mathcal{G}_r(\mathbf{F})_i =
\mathbf{F}[b_i,:,\lfloor v_i/r\rfloor,\lfloor u_i/r\rfloor].
\label{eq:bev_to_point}
\end{equation}
We concatenate the sampled features with the preceding descriptor of the same point and apply a shared point-wise MLP $\phi_k$ to obtain the updated descriptor $\mathbf{q}_i^{k}$:
\begin{equation}
\begin{aligned}
\widetilde{\mathbf{F}}^{k} &= \mathcal{B}_k(\mathbf{F}^{k-1}),\\
\mathbf{q}_i^{k} &=
\phi_k\bigl([\mathcal{G}_{r_k}(\widetilde{\mathbf{F}}^{k})_i,
                  \mathbf{q}_i^{k-1}]\bigr).
\end{aligned}
\label{eq:point_update}
\end{equation}
These descriptors form $\mathbf{Q}^{k}$, with the same channel width as $\widetilde{\mathbf{F}}^{k}$.

At stages $k=1,2,3$, we aggregate $\mathbf{Q}^{k}$ into a BEV tensor $\mathbf{S}^{k}$ for the next stage. At cumulative stride $r$, cell max pooling and scatter use the grouping indices $\mathcal{V}_i^{r}$ for point $i$:
\begin{equation}
\mathcal{V}_i^{r}=
\left(b_i,h_i,\left\lfloor v_i/r\right\rfloor,
                    \left\lfloor u_i/r\right\rfloor\right).
\label{eq:multiscale_cells}
\end{equation}
As $r$ increases, horizontal cells become coarser while each point's height partition $h_i$ stays fixed. Let $\mathcal{S}_r(\mathbf{Q})$ denote the resulting BEV tensor after arranging the pooled height-partition descriptors as channels. The aggregate is then $\mathbf{S}^{k}=\mathcal{S}_{r_k}(\mathbf{Q}^{k})$.

The current stage's BEV update combines the convolutional features $\widetilde{\mathbf{F}}^{k}$ with the incoming aggregate $\mathbf{S}^{k-1}$. We use bilinear interpolation ($\mathrm{Resize}$) to align $\mathbf{S}^{k-1}$ with the current BEV resolution, downsampling at stages 2--4. The convolutional fusion layer $\gamma_k$ projects their concatenation to the stage's BEV feature width (number of channels), producing $\mathbf{U}^{k}$. An attention module $\eta_k$ predicts sigmoid weights $\mathbf{A}^{k}$ that scale $\mathbf{U}^{k}$ before it is added to $\widetilde{\mathbf{F}}^{k}$. The stage's DSC module $\mathcal{D}_k$ then produces the updated BEV features $\mathbf{F}^{k}$. With $\odot$ denoting element-wise multiplication, these updates are
\begin{equation}
\begin{aligned}
\mathbf{U}^{k} &=
\gamma_k\bigl([\mathrm{Resize}(\mathbf{S}^{k-1}),
               \widetilde{\mathbf{F}}^{k}]\bigr),\\
\mathbf{A}^{k} &=
\eta_k(\mathbf{U}^{k}),\\
\mathbf{F}^{k} &=
\mathcal{D}_k\bigl(\widetilde{\mathbf{F}}^{k}
                   +\mathbf{A}^{k}\odot\mathbf{U}^{k}\bigr).
\end{aligned}
\label{eq:point_bev_fusion}
\end{equation}
The resulting BEV features $\mathbf{F}^{k}$ and point descriptors $\mathbf{Q}^{k}$ are retained for fusion across levels. Stages 1--3 also pass them, together with $\mathbf{S}^{k}$, to the next stage.

\subsubsection{\textbf{Dual-Scale BEV Context Module}}
\label{sec:dual_scale}

\begin{figure}[t]
\centering
\includegraphics[width=\columnwidth]{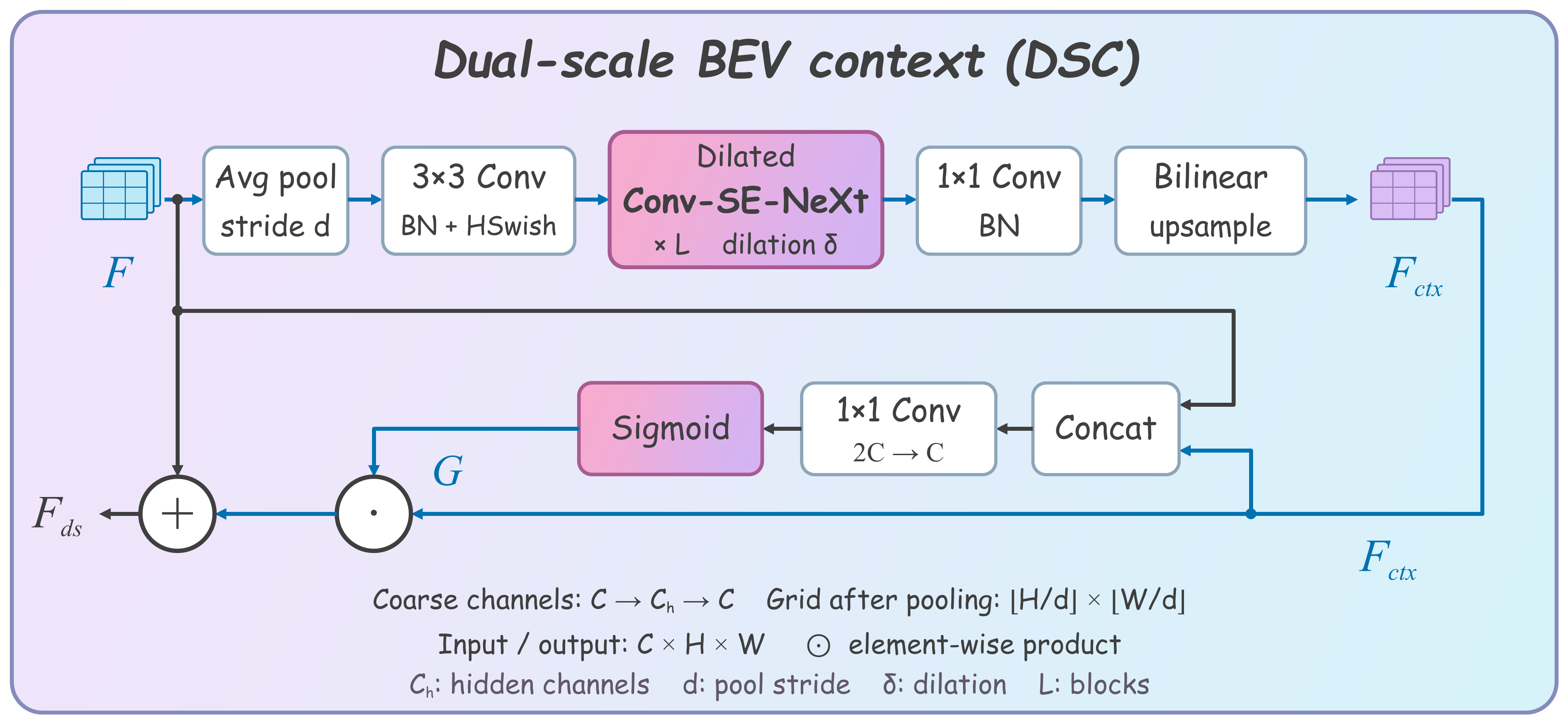}
\caption{Dual-scale BEV context (DSC) module.}
\label{fig:dsc_detail}
\end{figure}

Fig.~\ref{fig:dsc_detail} illustrates the DSC module $\mathcal{D}$, which adds context from a coarser BEV grid to the incoming features.

For input BEV features $\mathbf{F}$, we use $C$, $H$, and $W$ for the current input dimensions. Average pooling $\mathrm{Pool}_d$ with kernel size and stride $d$ produces a grid of $\lfloor H/d\rfloor\times\lfloor W/d\rfloor$ cells. A $3\times3$ convolution maps the pooled features from $C$ channels to hidden width $C_h$, followed by batch normalization and Hardswish. The features then pass through $L$ Conv-SE-NeXt blocks with dilation $\delta$ and stride one. A $1\times1$ convolution followed by batch normalization maps them back to $C$ channels. We denote this sequence of operations by $\kappa$. Bilinear interpolation ($\mathrm{Resize}_{H,W}$) upsamples the coarse features to the input resolution $H\times W$, yielding the context features $\mathbf{F}_{\mathrm{ctx}}$:
\begin{equation}
\mathbf{F}_{\mathrm{ctx}} =
\mathrm{Resize}_{H,W}\bigl(\kappa(\mathrm{Pool}_d(\mathbf{F}))\bigr),
\label{eq:coarse_context}
\end{equation}

We concatenate the input $\mathbf{F}$ with $\mathbf{F}_{\mathrm{ctx}}$ and apply a $1\times1$ convolution from $2C$ to $C$ channels, followed by the sigmoid function $\sigma$, to obtain a gate $\mathbf{G}$. The gate weighs the context at each spatial location and channel before it is added to the input, yielding the DSC output $\mathbf{F}_{\mathrm{ds}}$ with the same dimensions as the input:
\begin{equation}
\begin{aligned}
\mathbf{G} &= \sigma\left(\mathrm{Conv}_{1\times1}
\left([\mathbf{F},\mathbf{F}_{\mathrm{ctx}}]\right)\right),\\
\mathbf{F}_{\mathrm{ds}} &= \mathcal{D}(\mathbf{F})
= \mathbf{F}+\mathbf{G}\odot\mathbf{F}_{\mathrm{ctx}}.
\end{aligned}
\label{eq:dual_scale}
\end{equation}

\subsubsection{\textbf{Fuse Levels 0--4}}
We combine the stem and four stage outputs $\{\mathbf{F}^{j},\mathbf{Q}^{j}\}_{j=0}^{4}$ within each branch. The coarser BEV feature maps are upsampled to the stem resolution $H\times W$ by bilinear interpolation. We then concatenate all five feature maps along the channel dimension and apply 2D convolutions to obtain the fused BEV features $\mathbf{F}_{\mathrm{dec}}$. The point descriptors remain aligned with the same $N$ points across levels. We concatenate each point's descriptors from these levels and apply a shared point-wise MLP to obtain the fused point descriptors $\mathbf{Q}_{\mathrm{dec}}=\{\mathbf{q}_{\mathrm{dec},i}\}_{i=1}^{N}$.

\subsection{Point--BEV Fusion Head}
\label{sec:prediction}
The fusion head combines the backbone outputs $\mathbf{F}_{\mathrm{dec}}$ and $\mathbf{Q}_{\mathrm{dec}}$ for point-wise prediction. We sample the BEV features at each point using $\mathcal{G}_1$. A shared point-wise MLP $\mu$ maps these sampled features to the same width as $\mathbf{q}_{\mathrm{dec},i}$, and their sum forms the final descriptor $\mathbf{d}_i$. Let $K$ be the number of semantic classes. Dropout followed by a linear classifier produces the class-logit vector $\mathbf{o}_i\in\mathbb{R}^{K}$ for point $i$:
\begin{equation}
\begin{aligned}
\mathbf{d}_i &= \mu(\mathcal{G}_1(\mathbf{F}_{\mathrm{dec}})_i)
                +\mathbf{q}_{\mathrm{dec},i},\\
\mathbf{o}_i &= \mathrm{Linear}(\mathrm{Dropout}(\mathbf{d}_i)).
\end{aligned}
\label{eq:point_prediction}
\end{equation}
Collecting these vectors gives $\mathbf{O}\in\mathbb{R}^{N\times K}$; class-wise argmax gives the predicted label for each retained point.

\subsection{Auxiliary Supervision and Training Objective}
\label{sec:objective}

During training, auxiliary heads branch from the encoder and backbone to supervise intermediate features alongside the final point prediction. Each auxiliary classifier uses dropout followed by a convolution.

\subsubsection{\textbf{Shared 2D Auxiliary Head}}
\label{sec:encoder_supervision}
The shared head receives the encoder tensor $\mathbf{X}$. The same 2D classifier processes the features of the 3D cells in each height partition, producing logits $\mathbf{O}^{\mathrm{ht}}\in\mathbb{R}^{B\times Z\times K\times H\times W}$.

Each prediction is supervised by labels from its corresponding 3D cell. For valid ground-truth point labels $y_i\in\{0,\dots,K-1\}$, the target $\mathbf{Y}^{3\mathrm{D}}_{b,h,v,u}$ is the most frequent valid label among points in that cell. Cells without valid labels are ignored.

\subsubsection{\textbf{Stage Auxiliary Heads}}
The three stage heads receive the BEV maps $\mathbf{F}^{2}$, $\mathbf{F}^{3}$, and $\mathbf{F}^{4}$ (Fig.~\ref{fig:overall}). Each BEV feature map is upsampled to the stem resolution $H\times W$ by bilinear interpolation and passed to its classifier, producing logits $\mathbf{O}^{a}\in\mathbb{R}^{B\times K\times H\times W}$ for $a\in\{2,3,4\}$. All three heads use the targets $\mathbf{Y}^{2\mathrm{D}}\in\mathbb{Z}^{B\times H\times W}$, obtained by majority voting over all valid point labels in each horizontal cell across heights. Horizontal cells without valid labels are ignored.

\subsubsection{\textbf{Total Loss}}
The total objective $\mathcal{L}$ combines the point segmentation loss $\mathcal{L}_{\mathrm{seg}}$ and the auxiliary loss $\mathcal{L}_{\mathrm{aux}}$, with $\lambda$ weighting auxiliary supervision:
\begin{equation}
\mathcal{L}
=
\mathcal{L}_{\mathrm{seg}}
+
\lambda\mathcal{L}_{\mathrm{aux}} .
\label{eq:total_loss}
\end{equation}

The point segmentation loss supervises the fusion head logits $\mathbf{O}$ using the ground-truth point labels $\mathbf{y}=(y_1,\dots,y_N)$. It combines cross-entropy $\mathrm{CE}_{\mathbf{w},\epsilon}$ with the Lov\'asz--Softmax loss $\mathcal{L}_{\mathrm{Lovasz}}$~\cite{berman2018lovasz}. Here cross-entropy uses class weights $\mathbf{w}$ (unit weights in all experiments) and label smoothing $\epsilon$. Ground-truth labels designated as ignored by the dataset protocol do not contribute to the loss. $\mathrm{Softmax}$ acts along the class dimension:
\begin{equation}
\mathcal{L}_{\mathrm{seg}}
=
\mathrm{CE}_{\mathbf{w},\epsilon}(\mathbf{O},\mathbf{y})
+
0.3\,\mathcal{L}_{\mathrm{Lovasz}}
\left(
\mathrm{Softmax}(\mathbf{O}),\mathbf{y}
\right).
\label{eq:main_loss}
\end{equation}

We compute one loss for the shared 2D auxiliary head using the predictions and corresponding 3D-cell targets from all $Z$ height partitions of each of the $B$ scans. Each of the three stage auxiliary heads has a separate loss. Let $\ell_a$ denote the loss of head $a\in\{\mathrm{ht},2,3,4\}$. Their arithmetic mean over heads with valid targets gives $\mathcal{L}_{\mathrm{aux}}$; if no head has valid targets, this loss is zero.

For each auxiliary head $a$, let $\mathbf{O}^{a}$ and $\mathbf{Y}^{a}$ denote its logits and targets. The shared head uses $\mathbf{Y}^{\mathrm{ht}}=\mathbf{Y}^{3\mathrm{D}}$, while the stage heads use $\mathbf{Y}^{a}=\mathbf{Y}^{2\mathrm{D}}$.

Each auxiliary loss combines cross-entropy with Lov\'asz--Softmax and a boundary-F1-based loss $\mathcal{L}_{\mathrm{BD}}$~\cite{bokhovkin2019boundary}. The boundary loss is computed from the precision and recall of prediction--target boundary overlap. With coefficients $\alpha$ and $\beta$ weighting the auxiliary Lov\'asz--Softmax and boundary terms, respectively, the loss for head $a$ is
\begin{equation}
\begin{aligned}
\ell_a =
&\ \mathrm{CE}_{\mathbf{w},\epsilon}(\mathbf{O}^{a},\mathbf{Y}^{a})\\
&+\alpha\,
\mathcal{L}_{\mathrm{Lovasz}}
\left(\mathrm{Softmax}(\mathbf{O}^{a}),\mathbf{Y}^{a}\right)\\
&+\beta\,
\mathcal{L}_{\mathrm{BD}}
\left(\mathrm{Softmax}(\mathbf{O}^{a}),\mathbf{Y}^{a}\right).
\end{aligned}
\label{eq:aux_loss}
\end{equation}

\section{Experiments}

To assess whether HiPFuse can provide accurate semantics at the scan rate under limited computational resources, we measure segmentation accuracy (Sec.~\ref{sec:segmentation_results}) together with throughput and resource usage (Sec.~\ref{sec:efficiency}) on the outdoor benchmarks SemanticKITTI~\cite{behley2019semantickitti} and MCD~\cite{nguyen2024mcd}. Ablation studies (Sec.~\ref{sec:ablations}) examine how the proposed components contribute to segmentation accuracy, while separate training and evaluation on Semantic-ITC~\cite{wu2026semanticitc} assess indoor applicability (Sec.~\ref{sec:indoor_eval}). To determine whether these predictions benefit downstream robotic perception, we replace only the semantic input in fixed odometry, loop retrieval, and semantic mapping systems (Sec.~\ref{sec:downstream}).

\subsection{Experimental Setup}
\label{sec:experimental_setup}

\subsubsection{\textbf{Datasets}}
For MCD~\cite{nguyen2024mcd}, we use the NTU subset of outdoor campus scans acquired with a Livox Mid-70 LiDAR using a non-repetitive scan pattern. We use 24 learning classes and the published split of three training sequences and one test sequence. Following the released baselines, the test sequence is also used for checkpoint selection because no separate validation split is provided.

SemanticKITTI~\cite{behley2019semantickitti} contains urban road scans acquired with a Velodyne HDL-64E spinning LiDAR. We use 19 learning classes and the standard split of ten training, one validation, and eleven test sequences. Main comparisons use the official test set, while component and partition-count ablations use the validation sequence.

Semantic-ITC~\cite{wu2026semanticitc} contains indoor building scans acquired with a Livox Mid-360 LiDAR. We use 16 learning classes and its official sequence-level split of 37 training, 5 validation, and 10 test sequences.

In all segmentation benchmarks, the input consists only of coordinates and intensity from the current scan.

\subsubsection{\textbf{Spatial and Network Configurations}}
Table~\ref{tab:network_configs} summarizes the dataset-specific spatial and backbone settings. All three datasets use a stride-one BEV resolution of $256\times256$. All DSC instances use pooling stride $d=2$ and dilation $\delta=2$, with $L=2$ after stem fusion and $L=1$ after each stage.

\begin{table}[t]
\caption{Key HiPFuse settings. BEV widths and Conv-SE-NeXt block counts follow stages 1--4.}
\label{tab:network_configs}
\centering
\footnotesize
\setlength{\tabcolsep}{2pt}
\begin{tabular}{@{}lccc@{}}
\toprule
Setting & MCD & SemanticKITTI & Semantic-ITC \\
\midrule
$z$ range (m) & $[-4.5,24.0)$ & $[-4.0,2.5)$ & $[-4.0,8.0)$ \\
Cell size $(s_x,s_y)$ (m) & $0.6\times0.4$ & $0.35\times0.35$ & $0.1875\times0.1875$ \\
Height partitions $Z$ & 6 & 3 & 8 \\
BEV widths (1--4) & 64/128/256/256 & 128/128/128/128 & 64/128/256/256 \\
Blocks (1--4) & 2/2/2/2 & 1/1/1/1 & 2/2/2/2 \\
\bottomrule
\end{tabular}
\end{table}

\subsubsection{\textbf{Training Protocol}}
We train our models on a single NVIDIA L40 GPU with mixed precision, using AdamW~\cite{loshchilov2019decoupled} and a cosine learning-rate schedule with warmup. For MCD, SemanticKITTI, and Semantic-ITC, respectively, batch sizes are 8, 4, and 2; initial learning rates are $8\times10^{-4}$, $10^{-3}$, and $8\times10^{-4}$. The maximum numbers of training epochs are 300, 100, and 150, respectively.

The same early-stopping rule is used across all three datasets: after at least 30 epochs, training stops if the mIoU used for checkpoint selection does not improve for ten consecutive epochs. Checkpoint selection uses MCD \texttt{ntu\_day\_10}, SemanticKITTI sequence 08, and the official Semantic-ITC validation split, respectively.

\subsubsection{\textbf{Augmentation and Losses}}
We augment training scans on all three datasets with flipping, rotation, scaling, and translation; MCD and Semantic-ITC additionally use point dropout and jitter, while SemanticKITTI uses instance CutMix~\cite{puy2023using}. The auxiliary loss weight is $\lambda=0.2$ on MCD and Semantic-ITC and $\lambda=1.0$ on SemanticKITTI; only SemanticKITTI adds auxiliary Lov\'asz--Softmax~\cite{berman2018lovasz} and boundary-F1-based terms~\cite{bokhovkin2019boundary}, with $\alpha=1.5$ and $\beta=1.0$, respectively.

\subsubsection{\textbf{Evaluation Protocol}}
For the main comparisons, we train our models with a fixed random seed of 0 for reproducibility. For baselines evaluated locally, we use the provided implementations and their default prediction procedures.

We evaluate segmentation accuracy using FP32 inference with batch size one, without test-time augmentation or point subsampling. Following WaffleIron~\cite{puy2023using}, we assign each point outside the predefined 3D region the predicted label of its nearest retained point in the same scan under 3D Euclidean distance. We report class-wise intersection over union (IoU) and mIoU over all valid ground-truth points.

\begin{table*}[!t]
\caption{Class-wise IoU and mIoU (\%) on the MCD NTU test sequence. Bold: best; underline: second best.}
\label{tab:mcd_main}
\centering
\scriptsize
\setlength{\tabcolsep}{2.2pt}
\resizebox{\textwidth}{!}{%
\begin{tabular}{l l c *{24}{r} c}
\toprule
Method & Category & Year &
\rotatebox{60}{Barrier} &
\rotatebox{60}{Bike} &
\rotatebox{60}{Building} &
\rotatebox{60}{Chair} &
\rotatebox{60}{Cliff} &
\rotatebox{60}{Container} &
\rotatebox{60}{Curb} &
\rotatebox{60}{Fence} &
\rotatebox{60}{Hydrant} &
\rotatebox{60}{Sign} &
\rotatebox{60}{Lanemark} &
\rotatebox{60}{Other} &
\rotatebox{60}{Parkinglot} &
\rotatebox{60}{Pedestrian} &
\rotatebox{60}{Pole} &
\rotatebox{60}{Road} &
\rotatebox{60}{Shelter} &
\rotatebox{60}{Sidewalk} &
\rotatebox{60}{Stairs} &
\rotatebox{60}{Structure-O} &
\rotatebox{60}{Traffic-C} &
\rotatebox{60}{Trunk} &
\rotatebox{60}{Vegetation} &
\rotatebox{60}{Vehicle} &
\rotatebox{60}{mIoU (\%)} \\
\midrule
SalsaNext~\cite{cortinhal2020salsanext} & Projection-based & 2020 &
49.6 & 0.2 & 81.3 & 37.2 & 6.4 & 0.0 & 48.7 & 36.7 & 9.3 & 35.5 &
45.7 & 30.2 & 12.2 & 11.8 & 27.5 & 91.4 & 43.7 & 66.3 & 0.0 & 0.0 &
8.1 & 31.1 & 84.0 & 48.2 & 33.5 \\
WaffleIron~\cite{puy2023using} & Projection-based & 2023 &
76.7 & 28.3 & 81.6 & 43.9 & 6.9 & 6.9 & 57.2 & 37.7 & 14.7 & 57.9 &
53.5 & 11.7 & \underline{39.7} & \underline{52.5} & 56.7 & 91.5 & 67.3 & 77.0 & \underline{65.8} & \underline{42.2} &
12.4 & 59.0 & 88.2 & 75.9 & 50.2 \\
\addlinespace[2pt]
Cylinder3D~\cite{zhu2021cylindrical} & Voxel-based & 2021 &
70.5 & 1.4 & 90.1 & 42.3 & 9.2 & 19.5 & 43.4 & 31.6 & 16.2 & 65.4 &
48.5 & 1.4 & 13.6 & 41.1 & 65.1 & 90.5 & 60.7 & 58.7 & 41.3 & 0.6 &
10.3 & 57.2 & 91.6 & 72.1 & 43.5 \\
SDSeg3D~\cite{li2022self} & Voxel-based & 2022 &
72.2 & \textbf{57.9} & 82.2 & \underline{57.3} & 0.8 & 0.0 & 47.6 & 31.4 & 24.4 & 67.1 &
49.1 & 0.0 & 12.5 & 45.3 & 61.0 & 91.6 & 69.0 & 64.8 & 49.0 & 0.0 &
11.1 & 56.0 & 89.4 & 80.0 & 46.7 \\
S-Former~\cite{lai2023spherical} & Voxel-based & 2023 &
79.1 & 46.1 & 88.7 & 50.4 & 16.6 & 23.1 & \underline{61.2} & 46.7 & \underline{51.2} & \underline{75.4} &
\underline{55.1} & 10.4 & \textbf{45.1} & \textbf{66.6} & \underline{73.9} & \underline{93.6} & \underline{78.1} & 77.8 & \textbf{72.6} & 37.6 &
26.4 & \underline{63.3} & 92.3 & \underline{83.7} & \underline{59.0} \\
\addlinespace[2pt]
SPVCNN~\cite{tang2020searching} & Fusion-based & 2020 &
65.5 & \underline{56.1} & 81.4 & 44.9 & 8.6 & 16.6 & 50.6 & 37.0 & 22.0 & 65.8 &
49.7 & 1.3 & 28.2 & 46.3 & 58.1 & 91.7 & 62.9 & \textbf{93.8} & 49.1 & 6.1 &
8.9 & 51.3 & 89.5 & 71.4 & 48.2 \\
HARP-NeXt~\cite{abou2025harp} & Fusion-based & 2025 &
\underline{87.5} & 1.2 & \underline{90.8} & 49.6 & \underline{16.8} & \underline{23.3} & 56.3 & \underline{52.1} & 26.7 & 66.9 &
54.3 & \textbf{51.7} & 20.1 & 20.8 & 60.3 & 93.1 & 64.0 & 75.2 & 20.2 & 40.2 &
\underline{35.6} & 59.0 & \underline{92.5} & 55.0 & 50.6 \\
Ours & Fusion-based & 2026 &
\textbf{93.9} & 2.4 & \textbf{96.4} & \textbf{59.7} & \textbf{27.1} & \textbf{23.5} & \textbf{61.3} & \textbf{74.0} & \textbf{62.1} & \textbf{80.3} &
\textbf{55.3} & \underline{51.4} & 27.4 & 32.8 & \textbf{78.9} & \textbf{93.7} & \textbf{80.1} & \underline{78.9} & \textbf{72.6} & \textbf{59.4} &
\textbf{37.6} & \textbf{70.0} & \textbf{94.8} & \textbf{86.1} & \textbf{62.5} \\
\bottomrule
\end{tabular}
}
\par\vspace{1.5ex}
\caption{Class-wise IoU and mIoU (\%) on the SemanticKITTI test set. Bold: best; underline: second best.}
\label{tab:semantickitti_test}
\centering
\scriptsize
\setlength{\tabcolsep}{2.2pt}
\resizebox{\textwidth}{!}{%
\begin{tabular}{l l c *{19}{r} c}
\toprule
Method & Category & Year &
\rotatebox{60}{Car} &
\rotatebox{60}{Bicycle} &
\rotatebox{60}{Motorcycle} &
\rotatebox{60}{Truck} &
\rotatebox{60}{Other-veh.} &
\rotatebox{60}{Person} &
\rotatebox{60}{Bicyclist} &
\rotatebox{60}{Motorcyclist} &
\rotatebox{60}{Road} &
\rotatebox{60}{Parking} &
\rotatebox{60}{Sidewalk} &
\rotatebox{60}{Other-ground} &
\rotatebox{60}{Building} &
\rotatebox{60}{Fence} &
\rotatebox{60}{Vegetation} &
\rotatebox{60}{Trunk} &
\rotatebox{60}{Terrain} &
\rotatebox{60}{Pole} &
\rotatebox{60}{Traffic-sign} &
\rotatebox{60}{mIoU (\%)} \\
\midrule
SalsaNext~\cite{cortinhal2020salsanext} & Projection-based & 2020 &
91.9 & 48.3 & 38.6 & 38.9 & 31.9 & 60.2 & 59.0 & 19.4 & \underline{91.7} & 63.7 &
75.8 & 29.1 & 90.2 & 64.2 & 81.8 & 63.6 & 66.5 & 54.3 & 62.1 & 59.5 \\
PolarNet~\cite{zhang2020polarnet} & Projection-based & 2020 &
93.8 & 40.3 & 30.1 & 22.9 & 28.5 & 43.2 & 40.2 & 5.6 & 90.8 & 61.7 &
74.4 & 21.7 & 90.0 & 61.3 & 84.0 & 65.5 & 67.8 & 51.8 & 57.5 & 54.3 \\
WaffleIron~\cite{puy2023using} & Projection-based & 2023 &
\underline{97.2} & \textbf{70.0} & \textbf{69.8} & 40.4 & \textbf{59.6} & \textbf{77.1} &
\textbf{75.5} & \underline{41.5} & 90.6 & \textbf{70.4} & \underline{76.4} & \textbf{38.9} &
\textbf{93.5} & \textbf{72.3} & \textbf{86.7} & \textbf{75.7} & \textbf{71.7} &
\textbf{66.2} & \textbf{71.9} & \textbf{70.8} \\
RangeRet~\cite{mosco2026revisiting} & Projection-based & 2026 &
93.5 & 50.6 & 55.4 & \underline{51.3} & 52.9 & 67.3 & 66.3 & 27.3 & 90.3 & 68.5 &
73.5 & 32.0 & 90.1 & 63.3 & 84.5 & 65.6 & 70.4 & 56.0 & 65.8 & 64.5 \\
\addlinespace[2pt]
Cylinder3D~\cite{zhu2021cylindrical} & Voxel-based & 2021 &
97.1 & \underline{67.6} & \underline{64.0} & 50.8 & 58.6 & \underline{73.9} & 67.9 & 36.0 & 91.4 & 65.1 &
75.5 & \underline{32.3} & 91.0 & 66.5 & 85.4 & 71.8 & 68.5 & 62.6 & 65.6 & \underline{67.8} \\
\addlinespace[2pt]
SPVNAS~\cite{tang2020searching} & Fusion-based & 2020 &
\textbf{97.3} & 51.5 & 50.8 & \textbf{59.8} & \underline{58.8} & 65.7 & 65.2 &
\textbf{43.7} & 90.2 & 67.6 & 75.2 & 16.9 & 91.3 & 65.9 & \underline{86.1} & \underline{73.4} &
\underline{71.0} & \underline{64.2} & \underline{66.9} & 66.4 \\
HARP-NeXt~\cite{abou2025harp} & Fusion-based & 2025 &
95.3 & 53.6 & 48.3 & 41.8 & 39.5 & 59.4 & 61.6 & 0.8 &
90.1 & 68.2 & 74.0 & 25.7 & 89.6 & 63.8 & 84.0 & 68.4 &
68.4 & 57.2 & 63.6 & 60.7 \\
Ours & Fusion-based & 2026 &
95.5 & 63.4 & 57.9 & 37.6 & 40.5 & 60.5 & \underline{68.5} & 16.9 &
\textbf{92.1} & \underline{69.1} & \textbf{77.0} & 26.6 & \underline{92.6} & \underline{68.0} & 84.6 &
68.5 & 70.2 & 61.5 & 58.8 & 63.7 \\
\bottomrule
\end{tabular}
}
\end{table*}

\subsection{Semantic Segmentation Results}
\label{sec:segmentation_results}

\subsubsection{\textbf{MCD}}

Our method leads the MCD~\cite{nguyen2024mcd} comparison with 62.5\% mIoU (Table~\ref{tab:mcd_main}) and achieves the highest reported IoU in 19 of the 24 classes, including a tie on stairs. The largest margins over the next-best results occur on fence and other structures, while bike and pedestrian remain well below the best compared results.

Class-wise results for SalsaNext~\cite{cortinhal2020salsanext}, SPVCNN~\cite{tang2020searching}, Cylinder3D~\cite{zhu2021cylindrical}, SDSeg3D~\cite{li2022self}, WaffleIron~\cite{puy2023using}, and S-Former~\cite{lai2023spherical} are taken from the MCD dataset paper~\cite{nguyen2024mcd}. We train HARP-NeXt~\cite{abou2025harp} using the same MCD training protocol and hyperparameters as our method and report its seed-0 result.

\subsubsection{\textbf{SemanticKITTI}}

Our method reaches 63.7\% test mIoU on SemanticKITTI~\cite{behley2019semantickitti} (Table~\ref{tab:semantickitti_test}), with the highest IoUs for road and sidewalk and the second-highest IoUs for parking, building, fence, and bicyclist in this comparison. Improvements over HARP-NeXt~\cite{abou2025harp} cover 17 of the 19 classes; truck and traffic-sign are the two exceptions. Relative to WaffleIron~\cite{puy2023using}, larger deficits on motorcyclist, other-vehicle, and person outweigh the gains on road and sidewalk in the overall mIoU.

Results for SalsaNext~\cite{cortinhal2020salsanext}, WaffleIron, Cylinder3D~\cite{zhu2021cylindrical}, and SPVNAS~\cite{tang2020searching} are taken from WaffleIron~\cite{puy2023using}. Results for RangeRet~\cite{mosco2026revisiting} and PolarNet~\cite{zhang2020polarnet} are taken from their respective papers. Because HARP-NeXt~\cite{abou2025harp} reports only validation results, we use its released checkpoint and official code. Test results for both HARP-NeXt and our method are obtained from the official SemanticKITTI benchmark server under the single-scan evaluation protocol~\cite{behley2019semantickitti}.

\subsubsection{\textbf{Indoor Evaluation}}
\label{sec:indoor_eval}

On Semantic-ITC~\cite{wu2026semanticitc}, our method achieves 71.6\% mIoU (Table~\ref{tab:semantic_itc_main}), exceeding SalsaNext~\cite{cortinhal2020salsanext} but remaining below HARP-NeXt~\cite{abou2025harp} and WaffleIron~\cite{puy2023using}. It ranks second among the compared methods on floor, table, sofa, cabinet/shelf, and light. Compared with HARP-NeXt~\cite{abou2025harp}, it gains 14.8 percentage points on cabinet/shelf, while the deficits on stair and board account for most of its overall mIoU gap.

\begin{table*}[t]
\caption{Class-wise IoU and mIoU (\%) on the Semantic-ITC test set. Bold: best; underline: second best.}
\label{tab:semantic_itc_main}
\centering
\scriptsize
\setlength{\tabcolsep}{2.2pt}
\resizebox{\textwidth}{!}{%
\begin{tabular}{l l c *{16}{r} c}
\toprule
Method & Category & Year &
\rotatebox{60}{Wall} &
\rotatebox{60}{Floor} &
\rotatebox{60}{Ceiling} &
\rotatebox{60}{Column} &
\rotatebox{60}{Beam} &
\rotatebox{60}{Door} &
\rotatebox{60}{Stair} &
\rotatebox{60}{Table} &
\rotatebox{60}{Chair} &
\rotatebox{60}{Sofa} &
\rotatebox{60}{Cab./shelf} &
\rotatebox{60}{Screen} &
\rotatebox{60}{Light} &
\rotatebox{60}{Vegetation} &
\rotatebox{60}{Board} &
\rotatebox{60}{Other} &
\rotatebox{60}{mIoU (\%)} \\
\midrule
SalsaNext~\cite{cortinhal2020salsanext} & Projection-based & 2020 &
83.1 & 90.7 & 92.2 & 43.9 & 76.9 & 51.0 & 47.3 & 76.9 & 69.6 & 52.7 &
57.9 & \underline{91.2} & 90.0 & 89.8 & 31.5 & 72.2 & 69.8 \\
WaffleIron~\cite{puy2023using} & Projection-based & 2023 &
\underline{88.4} & \textbf{93.4} & \underline{93.8} & \textbf{55.7} & \textbf{81.0} & \textbf{60.7} &
\textbf{79.6} & \textbf{83.0} & \textbf{80.2} & \textbf{73.8} & \textbf{67.3} &
\textbf{93.8} & \textbf{97.2} & \textbf{94.7} & \underline{44.8} & \textbf{81.1} &
\textbf{79.3} \\
\addlinespace[2pt]
HARP-NeXt~\cite{abou2025harp} & Fusion-based & 2025 &
\textbf{88.5} & 92.1 & \textbf{94.4} & \underline{55.5} & \underline{80.1} &
\underline{56.1} & \underline{63.8} & 79.9 & \underline{75.6} & 68.2 &
43.2 & 86.7 & 94.5 & \underline{91.2} & \textbf{55.6} & \underline{77.9} &
\underline{75.2} \\
Ours & Fusion-based & 2026 &
86.7 & \underline{92.6} & 93.7 & 53.2 & 76.7 & 52.1 &
37.5 & \underline{81.7} & 69.8 & \underline{71.5} & \underline{58.0} &
85.9 & \underline{95.1} & 90.1 & 27.1 & 73.4 & 71.6 \\
\bottomrule
\end{tabular}
}
\end{table*}

\subsubsection{\textbf{Qualitative Comparison}}

\begin{figure*}[t]
\centering
\includegraphics[width=\textwidth]{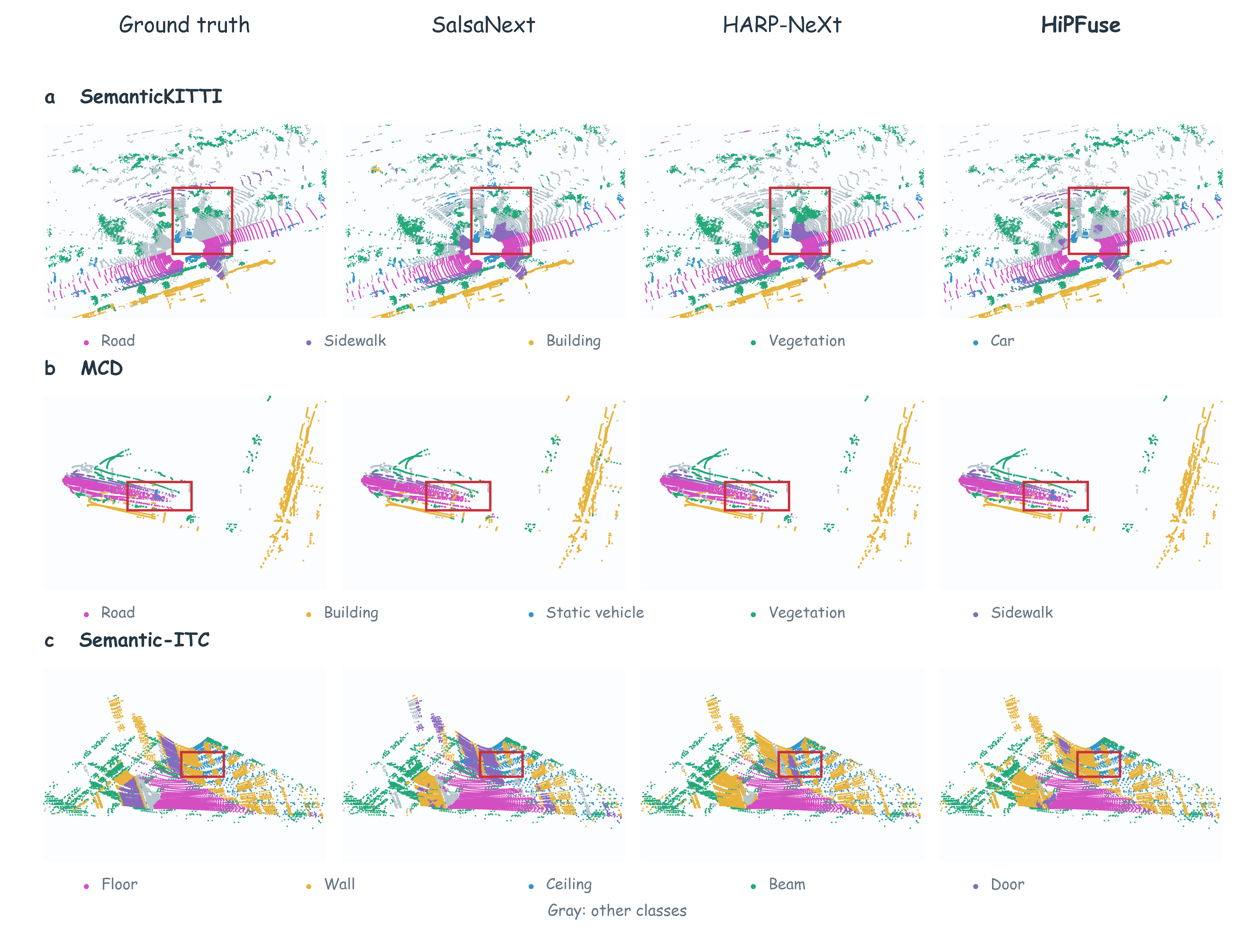}
\caption{Selected semantic predictions on (a) SemanticKITTI validation, (b) MCD test, and (c) Semantic-ITC test scans. Red boxes identify corresponding regions across methods. Row legends identify the colored classes; gray denotes other valid classes.}
\label{fig:semantic_qualitative}
\end{figure*}

Fig.~\ref{fig:semantic_qualitative} compares our predictions with those of SalsaNext~\cite{cortinhal2020salsanext} and HARP-NeXt~\cite{abou2025harp}, using the same points and view within each row. In the highlighted regions, our predictions more closely follow the reference road/sidewalk boundary on SemanticKITTI~\cite{behley2019semantickitti}, recover an MCD~\cite{nguyen2024mcd} vehicle largely classified as building by both baselines, and reduce wall-to-door confusion on Semantic-ITC~\cite{wu2026semanticitc}.

\subsection{Efficiency and Resource Analysis}
\label{sec:efficiency}

We complement the accuracy comparisons with segmentation throughput and peak GPU memory measurements on an NVIDIA L40 and a 64~GB Jetson AGX Orin (Table~\ref{tab:efficiency_comparison}). The Orin runs in MAXN mode with fixed CPU, GPU, and memory clock frequencies. Measurements cover the MCD~\cite{nguyen2024mcd} NTU test set, the SemanticKITTI~\cite{behley2019semantickitti} validation set, and the Semantic-ITC~\cite{wu2026semanticitc} test set. Each method uses the same configuration on both devices for a given dataset.

Timing spans scan-file input through recovery of CPU label arrays in original point order, including preprocessing, inference, and postprocessing. Time per scan is total traversal time divided by scan count; FPS is scan count divided by total traversal time in seconds. GPU memory is measured as peak active allocation in MiB.

\begin{table*}[t]
\caption{Segmentation throughput and resource usage on NVIDIA L40 and Orin. Bold and underline mark the highest and second-highest FPS per dataset and device.}
\label{tab:efficiency_comparison}
\centering
\small
\setlength{\tabcolsep}{3pt}
\resizebox{\textwidth}{!}{%
\begin{tabular}{c l l r r r r r r r}
\toprule
Dataset &
Method &
Category &
\shortstack{Params\\(M)} &
\multicolumn{3}{c}{NVIDIA L40 GPU} &
\multicolumn{3}{c}{Orin} \\
\cmidrule(lr){5-7}\cmidrule(lr){8-10}
&
&
&
&
\shortstack{GPU Memory\\(MiB)} &
\shortstack{Time/scan\\(ms)} &
\shortstack{FPS} &
\shortstack{GPU Memory\\(MiB)} &
\shortstack{Time/scan\\(ms)} &
\shortstack{FPS} \\
\midrule
\multirow{4}{*}{\rotatebox{90}{MCD~\cite{nguyen2024mcd}}} &
SalsaNext~\cite{cortinhal2020salsanext} &
Projection-based & 6.7 & 294.5 & 13.2 & \textbf{75.7} & 186.7 & 29.9 & \textbf{33.5} \\
&
SPVCNN~\cite{tang2020searching} &
Fusion-based & 21.8 & 146.1 & 77.8 & 12.8 & 151.2 & 61.5 & 16.3 \\
&
HARP-NeXt~\cite{abou2025harp} &
Fusion-based & 5.4 & 384.9 & 20.3 & 49.1 & 276.9 & 45.8 & 21.8 \\
&
Ours &
Fusion-based & 14.3 & 358.4 & 18.4 & \underline{54.3} & 359.4 & 44.6 & \underline{22.4} \\
\specialrule{0.8pt}{2pt}{2pt}
\multirow{8}{*}{\rotatebox{90}{SemanticKITTI~\cite{behley2019semantickitti}}} &
SalsaNext~\cite{cortinhal2020salsanext} &
Projection-based & 6.7 & 360.1 & 13.7 & \underline{72.9} & 1167.0 & 64.2 & \textbf{15.6} \\
&
PolarNet~\cite{zhang2020polarnet} &
Projection-based & 13.6 & 1473.6 & 66.0 & 15.1 & 1345.6 & 177.6 & 5.6 \\
&
WaffleIron~\cite{puy2023using} &
Projection-based & 6.8 & 1354.1 & 161.9 & 6.2 & 1235.1 & 529.1 & 1.9 \\
&
RangeRet~\cite{mosco2026revisiting} &
Projection-based & 3.8 & 760.6 & 32.5 & 30.8 & 761.8 & 177.9 & 5.6 \\
&
Cylinder3D~\cite{zhu2021cylindrical} &
Voxel-based & 55.9 & 1394.6 & 107.4 & 9.3 & 1394.6 & 448.9 & 2.2 \\
&
SPVNAS~\cite{tang2020searching} &
Fusion-based & 10.8 & 681.7 & 29.4 & 33.9 & 682.1 & 212.2 & 4.7 \\
&
HARP-NeXt~\cite{abou2025harp} &
Fusion-based & 5.4 & 938.5 & 19.8 & 50.4 & 856.0 & 118.3 & 8.5 \\
&
Ours &
Fusion-based & 7.2 & 659.1 & 12.6 & \textbf{79.0} & 660.1 & 70.5 & \underline{14.2} \\
\specialrule{0.8pt}{2pt}{2pt}
\multirow{4}{*}{\rotatebox{90}{\scriptsize\shortstack{Semantic-ITC\\\cite{wu2026semanticitc}}}} &
SalsaNext~\cite{cortinhal2020salsanext} &
Projection-based & 6.7 & 285.2 & 8.9 & \textbf{111.7} & 81.1 & 33.5 & \textbf{29.8} \\
&
WaffleIron~\cite{puy2023using} &
Projection-based & 6.8 & 462.4 & 166.0 & 6.0 & 256.3 & 133.7 & 7.5 \\
&
HARP-NeXt~\cite{abou2025harp} &
Fusion-based & 5.4 & 459.8 & 12.8 & 78.0 & 350.4 & 54.1 & 18.5 \\
&
Ours &
Fusion-based & 16.3 & 441.9 & 10.6 & \underline{93.9} & 442.9 & 50.9 & \underline{19.6} \\
\bottomrule
\end{tabular}
}
\end{table*}

On Orin, our method reaches 22.4~FPS on MCD and 14.2~FPS on SemanticKITTI, exceeding the 10~Hz scan rate on both outdoor benchmarks. It is more accurate but slower than SalsaNext~\cite{cortinhal2020salsanext} on these benchmarks. Relative to HARP-NeXt~\cite{abou2025harp}, it combines higher test mIoU and throughput with lower GPU memory use on SemanticKITTI; on MCD, higher mIoU accompanies similar throughput and greater GPU memory use. The separately trained indoor model reaches 19.6~FPS on Semantic-ITC, also exceeding the scan rate.

\subsection{Ablation Studies}
\label{sec:ablations}

\subsubsection{\textbf{Component Ablation}}
\label{sec:component_ablation}

All variants in Table~\ref{tab:ablation_components} retain the point--BEV backbone and three stage-level auxiliary heads. The two baselines compare single-layer and height-partitioned aggregation with relative-$z$ encoding, encoder-level supervision, and DSC disabled. Three further variants each remove one component: the two relative-$z$ offsets, the shared encoder-level auxiliary head, or the DSC modules. Removing the relative-$z$ offsets preserves the point encoder, raw coordinates, and horizontal offsets.

\begin{table*}[t]
\caption{Component ablation on MCD test and SemanticKITTI validation with fixed training settings per dataset. Full-model results summarize multiple training runs; other variants use seed 0.}
\label{tab:ablation_components}
\centering
\small
\setlength{\tabcolsep}{3pt}
\resizebox{\textwidth}{!}{%
\begin{tabular}{>{\raggedright\arraybackslash}p{0.21\textwidth} >{\raggedright\arraybackslash}p{0.13\textwidth} c c c c c}
\toprule
Variant &
Projection &
\shortstack{Relative-$z$\\encoding} &
\shortstack{Height-aligned\\BEV aux.} &
\shortstack{Dual-scale\\BEV context} &
\shortstack{mIoU on\\MCD (\%)} &
\shortstack{mIoU on\\SemanticKITTI (\%)} \\
\midrule
Single-layer point--BEV baseline &
Single-layer BEV &
\(\times\) & \(\times\) & \(\times\) &
60.6 & 64.3 \\
Height-partitioned point--BEV baseline &
Height-partitioned BEV &
\(\times\) & \(\times\) & \(\times\) &
60.5 & 63.2 \\
\addlinespace[1pt]
\cmidrule(lr){1-7}
\addlinespace[1pt]
w/o dual-scale BEV context &
Height-partitioned BEV &
\(\surd\) & \(\surd\) & \(\times\) &
\(60.9\) & \(64.6\) \\
w/o height-aligned BEV aux. &
Height-partitioned BEV &
\(\surd\) & \(\times\) & \(\surd\) &
\(60.4\) & \(64.0\) \\
w/o relative-$z$ offsets &
Height-partitioned BEV &
\(\times\) & \(\surd\) & \(\surd\) &
\(61.2\) & \(64.0\) \\
Full model &
Height-partitioned BEV &
\(\surd\) & \(\surd\) & \(\surd\) &
\(62.7 \pm 0.2\) & \(64.6 \pm 0.2\) \\
\bottomrule
\end{tabular}
}
\end{table*}

For the full model, Table~\ref{tab:ablation_components} reports the mean $\pm$ sample standard deviation of mIoU over training seeds 0, 1, and 2. The standard deviation is approximately 0.2 percentage points on both datasets, indicating consistent performance across these runs.

With relative-$z$ encoding, encoder-level supervision, and DSC disabled, height-partitioned aggregation does not improve on single-layer aggregation (Table~\ref{tab:ablation_components}). With these components retained, using $Z=6$ on MCD and $Z=3$ on SemanticKITTI improves mIoU over the single-layer setting ($Z=1$) by 0.7 and 0.9 percentage points, respectively (Table~\ref{tab:z_sensitivity}). These results support the value of height partitioning within the integrated HiPFuse framework. The full model achieves seed-0 mIoUs of 62.5\% on MCD and 64.8\% on SemanticKITTI, exceeding all five variants. Removing height-aligned supervision causes the largest decrease on MCD. On SemanticKITTI, removing either this supervision or the relative-$z$ offsets reduces mIoU to 64.0\%, while the result without DSC remains close to the full-model mean. Component effects therefore differ across datasets.

\subsubsection{\textbf{Sensitivity to Height Partitions}}

We vary the number of height partitions $Z$ while keeping all other network hyperparameters and training settings fixed (Table~\ref{tab:z_sensitivity}). The highest evaluated mIoU is obtained with $Z=6$ on MCD~\cite{nguyen2024mcd} and $Z=3$ on SemanticKITTI~\cite{behley2019semantickitti}, which are used in the corresponding models. The $Z=1$ setting retains the full model's encoding, auxiliary supervision, and DSC components. On MCD, $Z=6$ and $Z=8$ yield similar mIoU in this single-seed comparison. Since performance is not monotonic in $Z$, finer vertical partitioning is not uniformly beneficial.

\begin{table}[t]
\caption{Sensitivity to the number of height partitions $Z$ on MCD test and SemanticKITTI validation using seed 0. Bold marks the highest mIoU per dataset; -- denotes an unevaluated setting.}
\label{tab:z_sensitivity}
\centering
\small
\setlength{\tabcolsep}{5pt}
\begin{tabular}{c c c}
\toprule
\shortstack{Height\\partitions \(Z\)} &
\shortstack{mIoU on\\MCD (\%)} &
\shortstack{mIoU on\\SemanticKITTI (\%)} \\
\midrule
1 & \(61.8\) & \(63.9\) \\
3 & \(61.3\) & \(\mathbf{64.8}\) \\
6 & \(\mathbf{62.5}\) & \(63.7\) \\
8 & \(62.4\) & -- \\
\bottomrule
\end{tabular}
\end{table}

\subsection{Downstream Robotic Perception}
\label{sec:downstream}

HiPFuse and SalsaNext~\cite{cortinhal2020salsanext} are the only evaluated methods with measured segmentation throughput above the 10~Hz scan rate on Orin across all three datasets (Table~\ref{tab:efficiency_comparison}). We therefore compare their semantic inputs with those of HARP-NeXt~\cite{abou2025harp}, our baseline fusion architecture, in three downstream tasks. All downstream experiments run on Orin under the hardware and clock settings in Sec.~\ref{sec:efficiency}. Within each experiment, input scans, class mappings, downstream algorithms, and their parameters remain fixed while the semantic input changes.

All three tasks use SemanticKITTI~\cite{behley2019semantickitti} sequence 08 (4,071 scans). Odometry additionally uses MCD~\cite{nguyen2024mcd} sequence \texttt{ntu\_day\_10} (3,232 scans) and a separate 3,155-scan indoor route recorded with our handheld platform, equipped with a Livox Mid-360 and its built-in IMU. The indoor route traverses a campus building with environments similar to those in Semantic-ITC~\cite{wu2026semanticitc}. All runs process every scan in their respective evaluation sequences.

\subsubsection{\textbf{Semantic-Assisted Odometry}}
\label{sec:odometry}

Hybrid-VoxelMap~\cite{wu2025gaussian} uses planar/nonplanar semantic group assignments to select planar or Gaussian voxel models and their matching residuals. We compare the three learned semantic inputs with Gaussian-only mapping, and include ground-truth semantics for both outdoor sequences.

On SemanticKITTI, Hybrid-VoxelMap~\cite{wu2025gaussian} processes Velodyne LiDAR scans without IMU input and uses constant-velocity prediction. MCD uses Livox Mid-70 LiDAR and a VN100 IMU. We report outdoor odometry accuracy as the root mean square error (RMSE) of absolute trajectory error (ATE) after rigid SE(3) alignment.

In this real-world indoor experiment, we evaluate the models trained on Semantic-ITC~\cite{wu2026semanticitc} using our handheld recordings. The route starts and ends at the same physical location. Because trajectory and semantic ground truth are unavailable, we assess endpoint consistency using the endpoint position discrepancy, $e_{\mathrm{close}}=\lVert\mathbf{t}_{T}-\mathbf{t}_{1}\rVert_{2}$. Here $\mathbf{t}_{1}$ and $\mathbf{t}_{T}$ are the first and final estimated translations after initialization.

Semantic processing and SLAM run concurrently, following the original Hybrid-VoxelMap~\cite{wu2025gaussian} pipeline. We replay scans at processing speed while preserving their timestamps and IMU order. Odometry system throughput is the number of output poses divided by the elapsed time from the first scan publication to receipt of the final pose, including initialization. SLAM runtime is the mean active processing time per scan. We also report peak CPU memory usage for the SLAM process and peak GPU memory usage for the semantic front end.

\begin{figure*}[t]
\centering
\includegraphics[width=\textwidth]{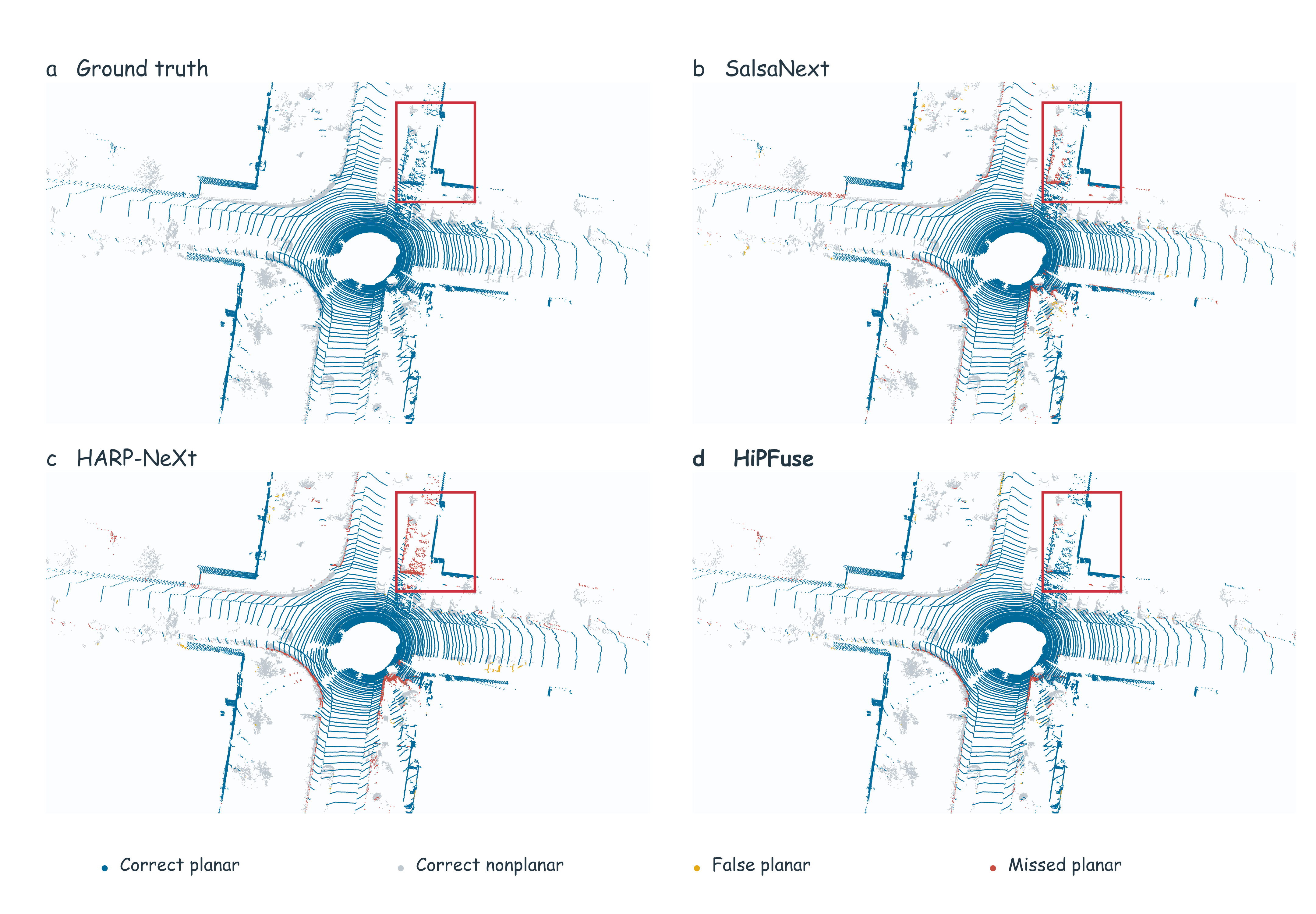}
\caption{Planar/nonplanar label grouping for Hybrid-VoxelMap~\cite{wu2025gaussian} on a selected SemanticKITTI 08 scan. Blue and gray indicate correct planar and nonplanar assignments; yellow and red indicate false-planar and missed-planar assignments, respectively. Red boxes mark corresponding regions. The reference groups are derived from semantic labels.}
\label{fig:planar_class_diff}
\end{figure*}

Fig.~\ref{fig:planar_class_diff} compares the predicted point-wise groups with those derived from ground-truth semantics. False-planar points are reference nonplanar points assigned to the planar group; missed-planar points are reference planar points assigned to the nonplanar group. Our method shows fewer assignments of both error types than SalsaNext~\cite{cortinhal2020salsanext} and HARP-NeXt~\cite{abou2025harp} in this scan. In the highlighted region, it retains more of the reference planar structure, preserving planar-model assignments that are lost in the baseline predictions.

\begin{table*}[t]
\caption{Outdoor semantic-assisted odometry with Hybrid-VoxelMap~\cite{wu2025gaussian} on Orin. Bold: lowest ATE among learned semantic inputs per dataset; --: inapplicable.}
\label{tab:semantic_slam}
\centering
\scriptsize
\setlength{\tabcolsep}{3pt}
\resizebox{\textwidth}{!}{%
\begin{tabular}{c l l r r r r r}
\toprule
Dataset &
Sequence &
Semantic input &
\multicolumn{3}{c}{SLAM} &
\multicolumn{1}{c}{Semantic front end} &
\multirow{2}{*}{\shortstack{System throughput\\(Hz) $\uparrow$}} \\
\cmidrule(lr){4-6}\cmidrule(lr){7-7}
&
&
&
\shortstack{ATE RMSE\\(m) $\downarrow$} &
\shortstack{Runtime\\(ms) $\downarrow$} &
\shortstack{CPU Memory\\(MiB) $\downarrow$} &
\shortstack{GPU Memory\\(MiB) $\downarrow$} & {} \\
\midrule
\multirow{5}{*}{SemanticKITTI~\cite{behley2019semantickitti}} &
08 & Gaussian-only, no semantics & 3.83 & 70.1 & 855.2 & -- & 14.1 \\
& 08 & Ground-truth semantics & 3.48 & 75.5 & 856.4 & -- & 13.1 \\
\noalign{\vskip 2pt}
\cdashline{2-8}[3pt/2pt]
\noalign{\vskip 2pt}
& 08 & SalsaNext~\cite{cortinhal2020salsanext} semantics & 3.77 & 81.8 & 852.2 & 184.4 & 11.2 \\
& 08 & HARP-NeXt~\cite{abou2025harp} semantics & 3.76 & 80.5 & 858.6 & 804.6 & 8.0 \\
& 08 & Our semantics & \textbf{3.71} & 84.4 & 852.4 & 626.9 & 11.1 \\
\specialrule{0.8pt}{2pt}{2pt}
\multirow{5}{*}{MCD~\cite{nguyen2024mcd}} &
\texttt{ntu\_day\_10} & Gaussian-only, no semantics & 2.44 & 37.2 & 771.3 & -- & 26.3 \\
& \texttt{ntu\_day\_10} & Ground-truth semantics & 1.65 & 38.0 & 772.6 & -- & 25.7 \\
\noalign{\vskip 2pt}
\cdashline{2-8}[3pt/2pt]
\noalign{\vskip 2pt}
& \texttt{ntu\_day\_10} & SalsaNext~\cite{cortinhal2020salsanext} semantics & 1.83 & 43.0 & 771.4 & 112.0 & 22.5 \\
& \texttt{ntu\_day\_10} & HARP-NeXt~\cite{abou2025harp} semantics & 2.17 & 44.2 & 772.1 & 231.6 & 19.4 \\
& \texttt{ntu\_day\_10} & Our semantics & \textbf{1.74} & 44.1 & 769.5 & 289.1 & 19.6 \\
\bottomrule
\end{tabular}
}
\end{table*}

All three learned semantic inputs reduce ATE relative to Gaussian-only mapping on both outdoor sequences, with larger reductions on MCD (Table~\ref{tab:semantic_slam}). Our semantic input yields the lowest ATE among the learned inputs on both sequences. The improvement over SalsaNext~\cite{cortinhal2020salsanext} and HARP-NeXt~\cite{abou2025harp} semantic inputs is modest on SemanticKITTI, with 3.71~m versus 3.76--3.77~m, and larger on MCD, with 1.74~m versus 1.83--2.17~m. Ground-truth semantics yield the lowest overall ATE, at 3.48 and 1.65~m, respectively.

With our semantic input, odometry system throughput reaches 11.1, 19.6, and 15.2~Hz on SemanticKITTI, MCD, and the indoor route, respectively. HARP-NeXt~\cite{abou2025harp} semantic input yields 8.0~Hz on SemanticKITTI, below the 10~Hz scan rate. Its throughput on MCD and the indoor route, 19.4 and 15.3~Hz, is close to that obtained with our semantic input.

\begin{table*}[t]
\caption{Indoor endpoint position discrepancy with Hybrid-VoxelMap~\cite{wu2025gaussian} on Orin. Bold: lowest discrepancy among learned semantic inputs; --: inapplicable.}
\label{tab:semantic_itc_slam}
\centering
\scriptsize
\setlength{\tabcolsep}{3pt}
\resizebox{\textwidth}{!}{%
\begin{tabular}{c l l r r r r r}
\toprule
Dataset &
Sequence &
Semantic input &
\multicolumn{3}{c}{SLAM} &
\multicolumn{1}{c}{Semantic front end} &
\multirow{2}{*}{\shortstack{System throughput\\(Hz) $\uparrow$}} \\
\cmidrule(lr){4-6}\cmidrule(lr){7-7}
&
&
&
\shortstack{Endpoint\\discrepancy (m) $\downarrow$} &
\shortstack{Runtime\\(ms) $\downarrow$} &
\shortstack{CPU Memory\\(MiB) $\downarrow$} &
\shortstack{GPU Memory\\(MiB) $\downarrow$} & {} \\
\midrule
\multirow{4}{*}{Indoor campus} &
Campus route & Gaussian-only, no semantics &
5.09 & 52.5 & 244.6 & -- & 18.8 \\
\noalign{\vskip 2pt}
\cdashline{2-8}[3pt/2pt]
\noalign{\vskip 2pt}
& Campus route & SalsaNext~\cite{cortinhal2020salsanext} semantics &
1.94 & 55.6 & 236.9 & 76.4 & 17.7 \\
& Campus route & HARP-NeXt~\cite{abou2025harp} semantics &
0.02 & 59.1 & 212.8 & 222.5 & 15.3 \\
& Campus route & Our semantics &
\textbf{0.01} & 59.1 & 217.7 & 364.0 & 15.2 \\
\bottomrule
\end{tabular}
}
\end{table*}

On the indoor route (Table~\ref{tab:semantic_itc_slam}), HARP-NeXt~\cite{abou2025harp} and HiPFuse yield endpoint discrepancies of 0.02~m and 0.01~m, respectively, versus 5.09~m without semantics and 1.94~m with SalsaNext~\cite{cortinhal2020salsanext}. Together with the outdoor ATE reductions, these results suggest that the benefit of semantic input varies with the scene. In Hybrid-VoxelMap~\cite{wu2025gaussian}, selecting planar models enables point-to-plane constraints on suitable surfaces, so their contribution depends on the geometry available for scan matching.

The Semantic-ITC~\cite{wu2026semanticitc} results in Table~\ref{tab:semantic_itc_main} show comparable IoUs for wall, floor, and ceiling, with most of our mIoU gap to HARP-NeXt arising from stair and board. All classes contribute equally to mIoU, while the recorded route contains few board objects and limited stair areas. The planar/nonplanar mapping assigns the same geometric model to several semantic classes, so confusions within a group leave model selection unchanged. The similarly small endpoint discrepancies are consistent with this class-wise accuracy pattern, the route's class distribution, and the planar/nonplanar grouping.

\subsubsection{\textbf{Loop Retrieval}}
\label{sec:loop_retrieval}

We evaluate loop retrieval with SGLC~\cite{wang2024sglc}. It builds foreground instance graphs from semantic labels, combines their spatial structure with background descriptors for retrieval, and verifies candidates through geometric matching between instances. Scans are replayed chronologically at a scheduled 10~Hz input rate, and semantic predictions are generated during replay. Each query accesses only previously processed scans, excludes the preceding 100 scans, and retains at most five candidates.

We also evaluate SegNet4D~\cite{wang2025segnet4d}, the default semantic segmentation method used by SGLC~\cite{wang2024sglc}. In our evaluation, it uses eight scans available up to the query time, aligned with KITTI ground-truth poses, whereas SalsaNext~\cite{cortinhal2020salsanext}, HARP-NeXt~\cite{abou2025harp}, and our method use single scans. We also evaluate retrieval with ground-truth semantic labels.

Among the 3,970 eligible queries, 321 have at least one eligible historical scan within 3~m, measured as the horizontal distance between their ground-truth positions. For these revisiting queries, Candidate Recall@5 measures the percentage with at least one such scan among the retained candidates. Accepted-loop recall measures the percentage for which such a scan is also accepted after geometric verification.

\begin{table}[t]
\centering
\caption{Loop retrieval with SGLC~\cite{wang2024sglc} on SemanticKITTI 08, measured on Orin. Bold: best among learned semantic inputs.}
\label{tab:sglc_online}
\footnotesize
\setlength{\tabcolsep}{3pt}
\begin{tabular}{lccc}
\toprule
Semantic input & \shortstack{Candidate\\Recall@5 (\%) $\uparrow$} & \shortstack{Accepted-loop\\recall (\%) $\uparrow$} & \shortstack{Semantic front-end\\service (ms) $\downarrow$} \\
\midrule
\shortstack[l]{Ground-truth\\semantics} & 93.2 & 70.4 & -- \\
\noalign{\vskip 2pt}
\hdashline[3pt/2pt]
\noalign{\vskip 2pt}
SalsaNext~\cite{cortinhal2020salsanext} & 81.6 & 57.0 & \textbf{92.8} \\
HARP-NeXt~\cite{abou2025harp} & 86.9 & 56.7 & 119.8 \\
SegNet4D~\cite{wang2025segnet4d}$^{\dagger}$ & 90.7 & 63.6 & 413.9 \\
Ours & \textbf{91.9} & \textbf{65.4} & 98.5 \\
\bottomrule
\end{tabular}
\par\smallskip
\begin{minipage}{\columnwidth}
\footnotesize
$^{\dagger}$Eight scans aligned using ground-truth poses.
\end{minipage}
\end{table}

With our semantic input, SGLC~\cite{wang2024sglc} achieves 91.9\% candidate recall and 65.4\% accepted-loop recall (Table~\ref{tab:sglc_online}), compared with 93.2\% and 70.4\% using ground-truth semantic input. Both measures exceed those obtained with all three learned baselines, including the multi-scan SegNet4D~\cite{wang2025segnet4d} reference.

Semantic front-end service time spans scan loading through point-label handoff to SGLC~\cite{wang2024sglc}, including preprocessing, inference, postprocessing, and label-file writing. The mean semantic front-end service times of our method and SalsaNext~\cite{cortinhal2020salsanext} are both below the 100~ms scan interval at 10~Hz.

\subsubsection{\textbf{Semantic Mapping}}
\label{sec:semantic_mapping}

We qualitatively compare semantic maps produced by SG-SLAM~\cite{wang2025sgslam} using the three learned semantic inputs and ground-truth labels. It uses semantic labels to organize and track object instances across scans, maintaining both semantic graph and point cloud maps. We use the default settings with loop closure enabled. Each map combines semantic point clouds from keyframes using the final poses estimated in that run.

\begin{figure*}[t]
\centering
\includegraphics[width=\textwidth]{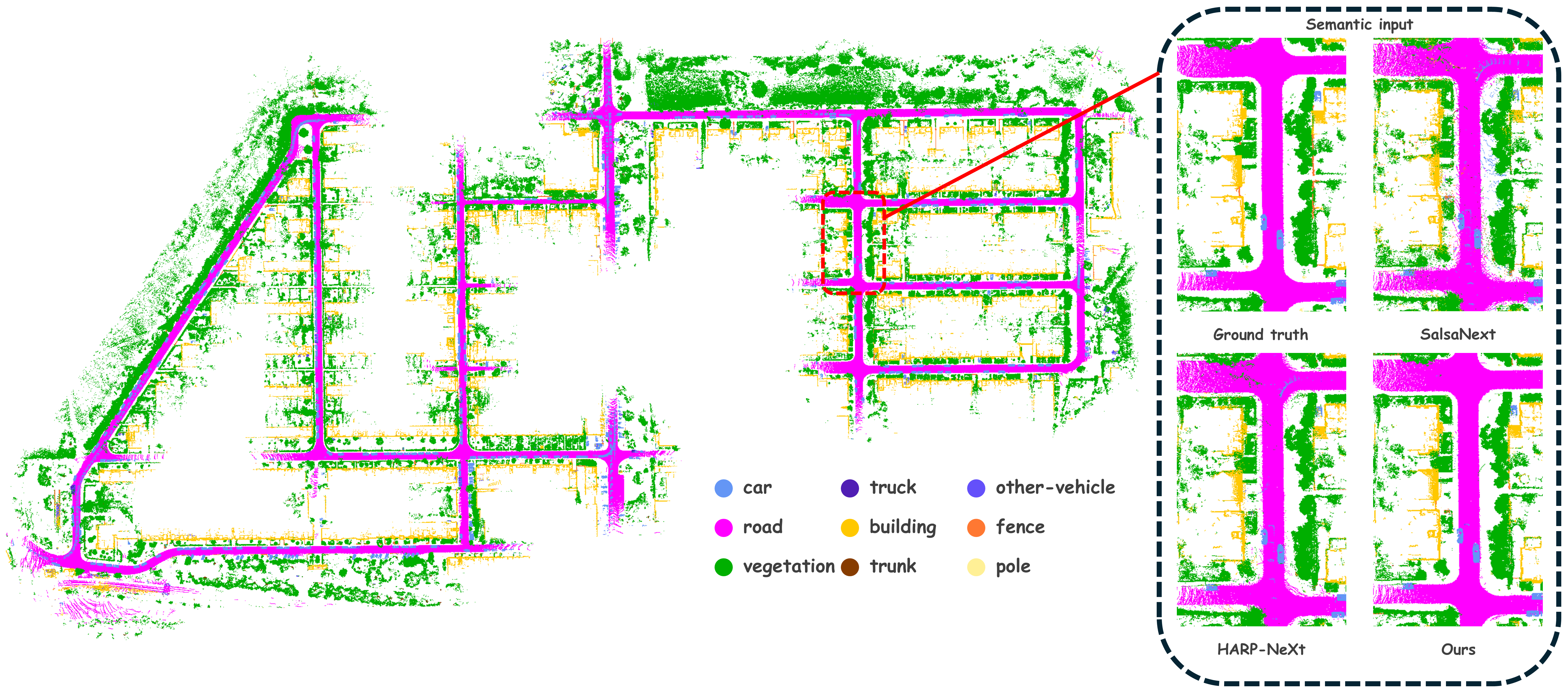}
\caption{Semantic maps generated with SG-SLAM~\cite{wang2025sgslam} on SemanticKITTI 08. Left: global map generated using our semantic input; the compared region is boxed in red. Right: local maps from four semantic inputs, each with its own estimated poses. Ground truth in the panel labels denotes the semantic labels used as input.}
\label{fig:semantic_mapping}
\end{figure*}

Around the two road junctions compared in Fig.~\ref{fig:semantic_mapping}, our result has fewer scattered non-road labels and a more continuous road-labeled surface than the results with SalsaNext~\cite{cortinhal2020salsanext} and HARP-NeXt~\cite{abou2025harp}. The road-label distribution also more closely resembles that obtained with ground-truth semantic input.

\section{Conclusion}

We presented HiPFuse, a point--BEV fusion framework for efficient LiDAR semantic segmentation from a single scan. Its integrated design retains height and point distinctions while enabling spatial context extraction through 2D processing for robotic perception.

The downstream evaluations show that HiPFuse's semantic predictions benefit odometry, loop retrieval, and semantic mapping. Together with the measured edge-device throughput, this supports improving semantic input as a practical route to strengthening robotic perception under limited computational resources. The additional indoor evaluation supports applicability after separate training, although our model remains less accurate than the strongest baselines on that dataset.

Like the downstream systems it serves, HiPFuse is configured per sensor setup. A limitation is that its spatial grid, including the height range, cell size, and number of height partitions, is chosen together with supervised training, so adapting it to a new setup requires labeled data and retraining. This motivates further work on adaptive height partitioning and, more broadly, on universal LiDAR segmentation models across datasets and sensors~\cite{liu2024multispace} that maintain scan-rate inference on edge devices.

\bibliographystyle{IEEEtran}
\bibliography{references}

\end{document}